\documentclass[letterpaper, 10 pt, conference]{ieeeconf}  

\IEEEoverridecommandlockouts                          
\usepackage{cite}
\usepackage{amsmath,amssymb,amsfonts}
\usepackage{algorithmic}
\usepackage{graphicx}
\usepackage{textcomp}
\usepackage{xcolor}
\usepackage{float}
\usepackage{booktabs}
\usepackage{multirow}
\usepackage{todonotes}
\usepackage{subcaption}
\usepackage{multirow}
\usepackage{graphicx}
\usepackage[colorlinks]{hyperref}
\hypersetup{colorlinks, linkcolor=red}

\newcommand{\argmin}[1]{\underset{#1}{\operatorname{argmin}}}
\DeclareMathOperator*{\argmax}{arg\,max}
   
\def\BibTeX{{\rm B\kern-.05em{\sc i\kern-.025em b}\kern-.08em
    T\kern-.1667em\lower.7ex\hbox{E}\kern-.125emX}}
\title{\LARGE \bf
A Multi-Hypothesis Approach to Pose Ambiguity \\ in Object-Based SLAM}

\author{Jiahui Fu, Qiangqiang Huang, Kevin Doherty, Yue Wang, and John J. Leonard
\thanks{All authors are with the MIT Computer Science and Artificial Intelligence Laboratory, Cambridge, MA 02139, USA.
        {\tt\{jiahuifu,hqq,kdoherty,yuewangx,jleonard\}@mit.edu}}%
\thanks{This work was supported by ONR MURI grant N00014-19-1-2571 and ONR grant N00014-18-1-2832.}}

\begin{document}

\maketitle
\thispagestyle{empty}
\pagestyle{empty}

\begin{abstract}
 In object-based Simultaneous Localization and Mapping (SLAM), 6D object poses offer a compact representation of landmark geometry useful for downstream planning and manipulation tasks. However, measurement ambiguity then arises as objects may possess complete or partial object shape symmetries (e.g., due to occlusion), making it difficult or impossible to generate a \emph{single} consistent object pose estimate. One idea is to generate multiple pose candidates to counteract measurement ambiguity. In this paper, we develop a novel approach that enables an object-based SLAM system to reason about multiple pose hypotheses for an object, and synthesize this locally ambiguous information into a globally consistent robot and landmark pose estimation formulation. In particular, we (1) present a learned pose estimation network that provides multiple hypotheses about the 6D pose of an object; (2) by treating the output of our network as components of a mixture model, we incorporate pose predictions into a SLAM system, which, over successive observations, recovers a globally consistent set of robot and object (landmark) pose estimates. We evaluate our approach on the popular YCB-Video Dataset and a simulated video featuring YCB objects. Experiments demonstrate that our approach is effective in improving the robustness of object-based SLAM in the face of object pose ambiguity.\footnote{Video: \href{https://youtu.be/E4sheabxWBI}{\tt{https://youtu.be/E4sheabxWBI}}}
\end{abstract}

\section{Introduction}
 Object-based Simultaneous Localization and Mapping (SLAM) incorporates higher-level object primitives into the localization and mapping process. Compared to traditional approaches utilizing low-level geometric features, it tends to improve its robustness to measurement noise and textureless scenes and bears the potential to provide a rich representation for downstream scene understanding, planning, and manipulation tasks \cite{rosen2021advances}. By encoding abundant robot-landmark geometric constraints  in  a  compact  form,  6D  object  poses  act as  a desirable type of object measurements to be leveraged~\cite{salas2013slam++}. When perceiving the world in terms of objects, robots could then readily integrate these 6D object pose measurements into the joint recovery of robot and landmark poses.

However, when there exists environment ambiguity due to complete or partial object (landmark) shape symmetries (e.g., from occlusion), the frontend may be misled to make false positive pose measurements. For instance, the frontend may throw out a random measurement of a mug's orientation when its handle is obscured. Such a pose measurement may incur considerable estimation errors in the backend if it is used to recover robot and landmark poses. 
\begin{figure}[]
    \centering
    \includegraphics[width=\linewidth]{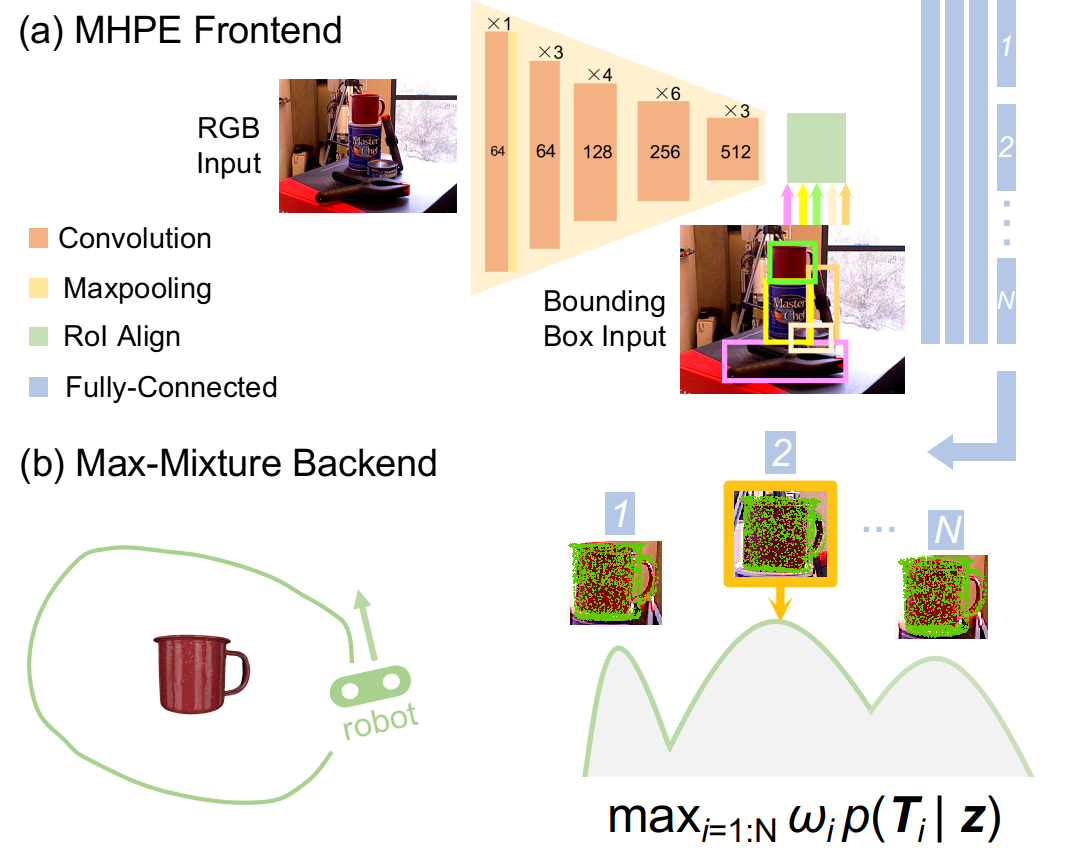}   
    \caption{Approach Overview. The system takes as input an RGB image along with its bounding box detection and outputs the optimized robot/landmark pose estimates. (a) The multi-hypothesis  6D  object  pose  estimator (MHPE) in the frontend comprises three components: feature extraction, region-of-Interest (RoI) Align, and a multilayer perceptron. The output is $N$ 6D pose hypotheses ($T_i$) for each object, coming from $N$ separate branches in the last layer (blue), which are then sent to the max-mixture backend. (b) The max-mixture backend models the multi-modality embedded in the multi-hypothesis output. With MHPE covering the correct object pose (orange box, shown in green points matching the mug) in a big fraction of the observed frames, max-mixtures could gradually converge to the dominant mode, i.e., correct estimates of robot/landmark poses.}
    \label{frame}
\end{figure}

One way to counteract the effect of ambiguity is to generate multiple hypotheses each time, in which a consistent set of hypothesis could be covered.  As a result, rather than conducting estimation based on one single frontend measurement at a time, as most current object-based SLAM systems do, we choose to enable the object-based SLAM system to be aware of the potential environment ambiguity through multiple hypotheses, i.e., generating and processing multiple possible object pose hypotheses for a single set of consistent estimates.  In this way, by considering multiple hypotheses across many observations, an object-based SLAM system could then employ past evidence to \emph{disambiguate} the true pose of the landmark (object) and recover a globally consistent set of robot and object pose estimates.

This idea motivates the design of our approach. Our objective is to enable an object-based  SLAM  system  to reason about multiple hypotheses for an object pose. By producing and synthesizing these locally  ambiguous  information  into  a joint multi-hypothesis robot/landmark pose estimation setting, the system could solve for a  globally  consistent set of object  and  robot  poses. Therefore, we instantiate our approach as follows 
(see Fig.~\ref{frame}): 
\begin{itemize}
\item To receive ambiguity-aware measurements from the frontend, we develop a deep-learning-based multi-hypothesis 6D object pose estimator (MHPE) trained with known 3D object models to explore the potential pose hypothesis space.
\item To solve the joint pose optimization with the multi-hypothesis formulation in the backend, we treat the multiple measurements as components  of  a  mixture  model and exploit nonlinear least square optimization to solve the max-mixture formulation~\cite{olson2013inference} of pose hypotheses.
\end{itemize}

Experiments are conducted on the real YCB-Video Dataset and a synthetic video sequence featuring YCB objects in a virtual environment. Results demonstrate our approach's effectiveness in improving  the  robustness  of  the object-based SLAM  system  in the face of ambiguous measurements.

Our paper proceeds as follows: we describe related work in Section II, elaborate on our pose estimation network design in Section III and max-mixtures formulation in Section IV, showcase experimental results in Section V, and conclude in Section VI. 

\section{Related Work}
\subsection{6D Object Pose Estimation}
Various approaches have been developed to address the object pose estimation problem with different data modalities. 

For RGB-based data inputs, traditional methods usually tackle the problem via feature matching between the query data and given object models~\cite{collet2011moped, zhu2014single,aubry2014seeing}, but are susceptible to low-texture regions or lighting changes. With the advancement in deep learning, PoseCNN~\cite{xiang2017posecnn} and SSD-6D~\cite{kehl2017ssd6d}, as two pioneering methods for estimating object pose in cluttered scenes, use neural networks to perform object detection and pose estimation at the same time. DOPE\cite{tremblay2018deep} uses only synthetic training data for robot manipulation tasks and the network shows good generalization to novel environments. Densefusion~\cite{wang2019densefusion} fuses pixel-wise dense features from RGB and depth images and achieves strong pose estimation results.

For point-cloud-based inputs, ICP~\cite{Besl1992} is the most classical and widely used approach though prone to local minima due to non-convexity. \cite{Hahnel02probabilisticmatching,agamennoniIROS16, iser_2016_hinzmann} use probabilistic approaches to improve resilience to uncertain data. For its learning-based variants, SegICP~\cite{Wong_2017} integrates semantic segmentation and pose estimation using neural networks, which achieves robust point cloud registration and per-pixel segmentation at the same time. The Deep Closest Point (DCP)~\cite{Wang_2019_ICCV} and Partial Registration Network (PRNet)~\cite{Wang_2019_NeurIPS} methods incorporate descriptor learning into the ICP pipeline, so that the model can learn matching priors from the data.

While all these above methods achieve good pose estimation accuracy, they either do not address the object shape ambiguity explicitly, or could not make a good compromise between estimation accuracy and processing speed, if serving as a SLAM frontend.

\subsection{Uncertainty-Aware SLAM Backend}
Uncertainty in SLAM involves many types of problems, such as loop closure and data association, where different backend methods have been developed. FastSLAM~\cite{1241885} tackles unknown data association in a particle-filter-based fashion that uses sampling to cover the overall probability space. Sünderhauf et al.~\cite{sunderhauf2012towards,6385590} develop methods to change the graph topology and achieve improved robustness to outliers with switchable constraints. The max-mixture method developed by Olson and Agrawal~\cite{olson2013inference} represents discrete hypotheses as components of max-mixtures; this approach has recently been leveraged for unknown data association in the context of semantic SLAM  by Doherty et al.~\cite{doherty2020probabilistic}.

\subsection{Object-Based SLAM}
SLAM++~\cite{salas2013slam++} uses pixel-wise reprojection error and adds 6D object poses as camera-object pose constraints to the pose-graph optimization. But it needs 3D object models at runtime and is limited to few categories of pre-selected objects. QuadricSLAM~\cite{nicholson2018quadricslam} manages to estimate camera poses and quadric parameters together based on odometry and bounding box detections. CubeSLAM~\cite{yang2019cubeslam} conducts single-view 3D cuboid object detection and incorporates its refinement into the joint optimization of camera poses and objects. While these two works improve the camera pose estimation by including constraints from novel representations of the objects, they do not consider explicitly about the impact of shape ambiguity on these novel representations. PoseRBPF~\cite{deng2019poserbpf} employs a Rao-Blackwellized particle filter to track object poses across consecutive frames in the frontend, demonstrating robustness to object shape symmetry when a sufficient number of sampling particles are used, which, however, limits the computation speed.

With object (landmark) poses as measurements, our work here enhances the robustness of object-based SLAM in ambiguous environments by effectively generating multiple pose hypotheses from a learned pose estimation network and, different from what PoseRBPF does, then incorporates sequential observations into the backend factor graph for obtaining smoothed estimates of all robot/landmark poses.
 
\section{Frontend: Multi-Hypothesis 6D Object Pose Estimator (MHPE)}
The MHPE frontend takes in an RGB image together with its object bounding boxes (could come from any off-the-shelf real-time object detectors), and outputs $N$ hypotheses for the 6D pose of each object detected in the scene. 

The object pose is represented by rotation $\textbf{q}=(w,x,y,z)$ as a unit quaternion and 3D translation $\textbf{t}=(t_x,t_y,t_z)$, which are defined with respect to the current camera frame. As $\textbf{q}=(w,x,y,z)$ and $\textbf{q}=(-w,-x,-y,-z)$ are equivalent, we enforce $\textbf{q}$ to have only non-negative real part $w$.
\subsection{MHPE Network}
The network processes the input data in three stages: feature extraction, region-of-interest alignment (RoI Align), and a multilayer perceptron (MLP) for final pose estimation (see Fig.~\ref{frame}).

We use ResNet-34~\cite{he2016deep} without the last average pooling layer to get pixel level feature embeddings. The feature maps, as well as  corresponding bounding boxes, are fed into the RoI Align layer. First introduced by MaskRCNN~\cite{he2017mask}, this layer performs bilinear interpolation on the image feature embeddings and produces feature maps of the same shape for each bounding box region. The last MLP component is divided into two parts, one for rotation estimation and the other for translation. For $N$ pose hypotheses for each object, the output size of the last fully connected layer is $4N$ for rotation $(w,x,y,z)$ and $3N$ for translation $(t_x,t_y,t_z)$. We apply Softplus to ensure non-negativity for “$w$” values. The final unit quaternion is obtained by normalizing the four outputs from the rotation branch.
\subsection{Learning Objective for MHPE Outputs} \label{learning obj}
The MHPE network aims to minimize the average distance (ADD loss\cite{hinterstoisser2012model}) between the two object point sets, transformed by the predicted and groundtruth poses, respectively. The ADD loss is defined as: 
\begin{equation}
l_{ADD}=\frac{1}{|\mathcal{X}|}\sum_{\mathbf{x}\in\mathcal{X}}\|(\mathbf{\hat{R}x+\hat{t}})-(\mathbf{Rx+t})\|\label{loss}
\end{equation}
where $\mathbf{R}\in\mathrm{SO}(3)$, $\mathbf{t}\in\mathrm{R}^3$, $[\mathbf{R}|\mathbf{t}]$ and $[\mathbf{\hat{R}}|\mathbf{\hat{t}}]$ are the groundtruth and estimated object poses, and $\mathbf{x}$ is the 3D point in the object point sets $\mathcal{X}$ from the known 3D object model. This metric performs well for the common single-hypothesis network, but needs adaption for the MHPE case, as different treatments of the individual loss in the hypothesis set may lead to distinct optimization directions.

Here, our pose estimation network is designed in a “mixture-of-expert” fashion~\cite{guzman2012multiple}, where the $N$ hypotheses are made independently based on a shared set of feature maps. It is thus ideal to have each output branch acquire some specialty in certain output domains, hence helping improve the estimation accuracy and shed extra light on possible causes of pose ambiguity.

In this spirit,  we choose to apply \eqref{loss} in a winner-takes-all fashion, which is formulated as:
\begin{equation}
\begin{split}
&\hat{\mathbf{P}}_i=\left<\hat{\mathbf{P}}_i^{(1)},\dots,\hat{\mathbf{P}}_i^{(N)}\right>\\
&L_i\left(\hat{\mathbf{P}}_i\right)=\argmin{j\in[1,N]} ~ l_{ADD}(\hat{\mathbf{P}}_i^{(j)})     
\end{split}
\label{wta}
\end{equation}
where $\hat{\mathbf{P}}_i^{(j)}=[\mathbf{R}(\mathbf{q}^{(j)}_i)|\mathbf{t}_{i}^{(j)}]$ is the $j^{th}$ pose estimate in the output ensemble for the $i^{th}$ input and $\mathbf{R}$ is expressed as the rotation matrix from quaternion $\mathbf{q}$.
\begin{figure}
    \centering
    \includegraphics[width=\linewidth]{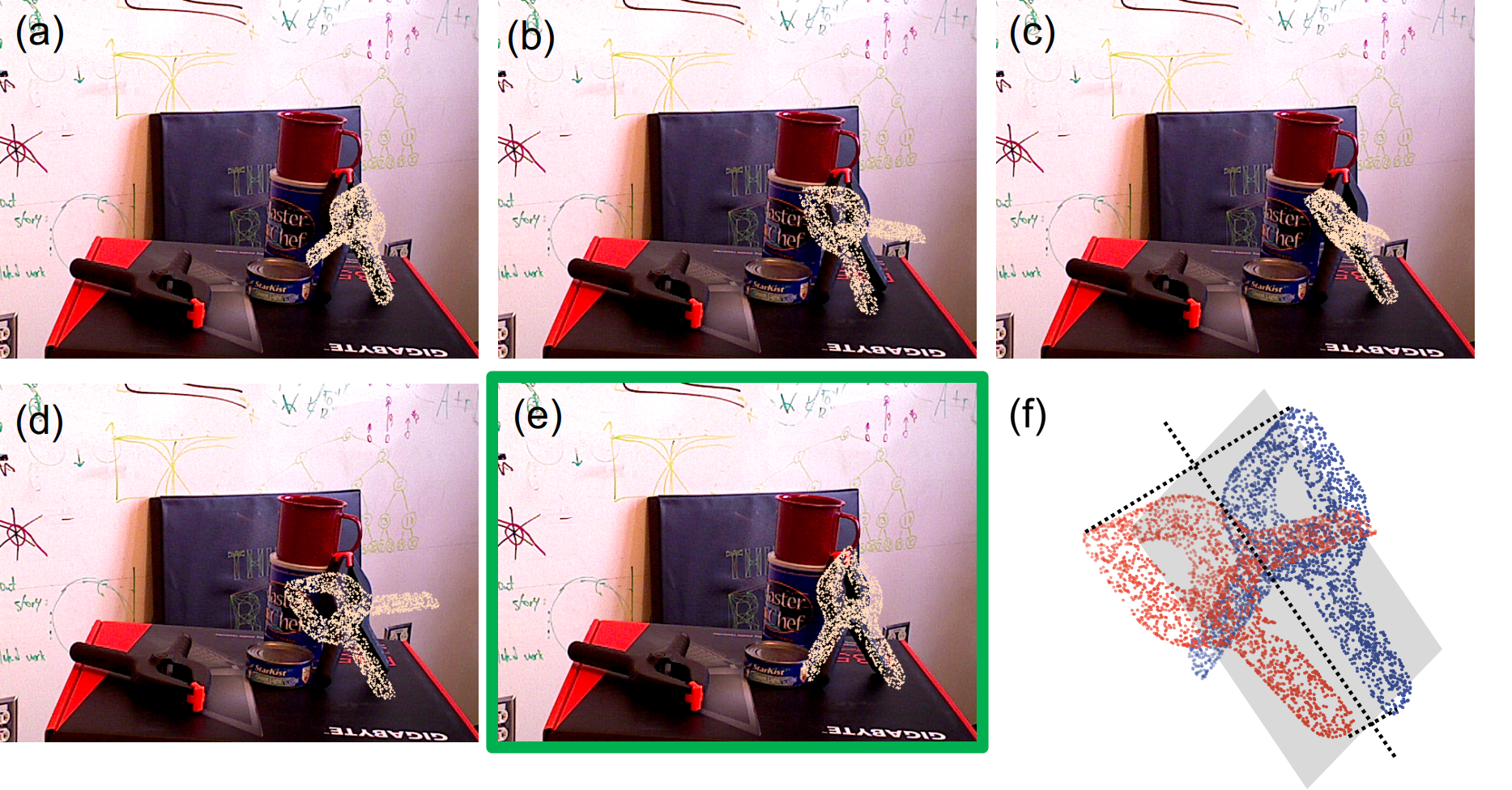}
    \caption{Hypothesis diversity from the MHPE frontend on the YCB object clamp: Hypotheses (a) \& (d) and (b) \& (e) are two quasi-mirror images; hypothesis (c) is spurious, and the green-boxed hypothesis (e) gives a relatively good estimate. (f) illustrates that the 3D point clouds of the mirror image pair could incur similar distance loss with respect to the grey mirror plane.}
    \label{variety}
\end{figure}
The defined loss function, $L_i$ for input $i$, only penalizes the most accurate estimation, regardless of “how bad” the rest of the hypotheses are. This treatment, compared to averaging, prevents other worse estimations from vanishing and adds some randomness, hence diversity, to the multi-branch estimator. Those branches with an initialization too far away from the minimum could thus slowly evolve to different domains of the output space, hedging their bets while refraining from paying the price of making the wrong tentative decision. Fig.~\ref{variety} shows one of our diverse pose estimation of a symmetrical clamp from the YCB-Video Dataset, illustrating the benefit of encouraging multiple hypotheses in different domains.

\section{Backend: Max-Mixture-Based Multi-Hypothesis Modeling and Optimization}
Compared to its common single-hypothesis counterpart, the MHPE frontend can better tackle the ambiguity within the current observation via giving multiple pose candidates. Nevertheless, this leads to exponential computation complexity for picking a pose hypothesis combination that will bring about a sequence of optimal pose estimates.

\begin{figure}
    \centering
    \includegraphics[width=\linewidth]{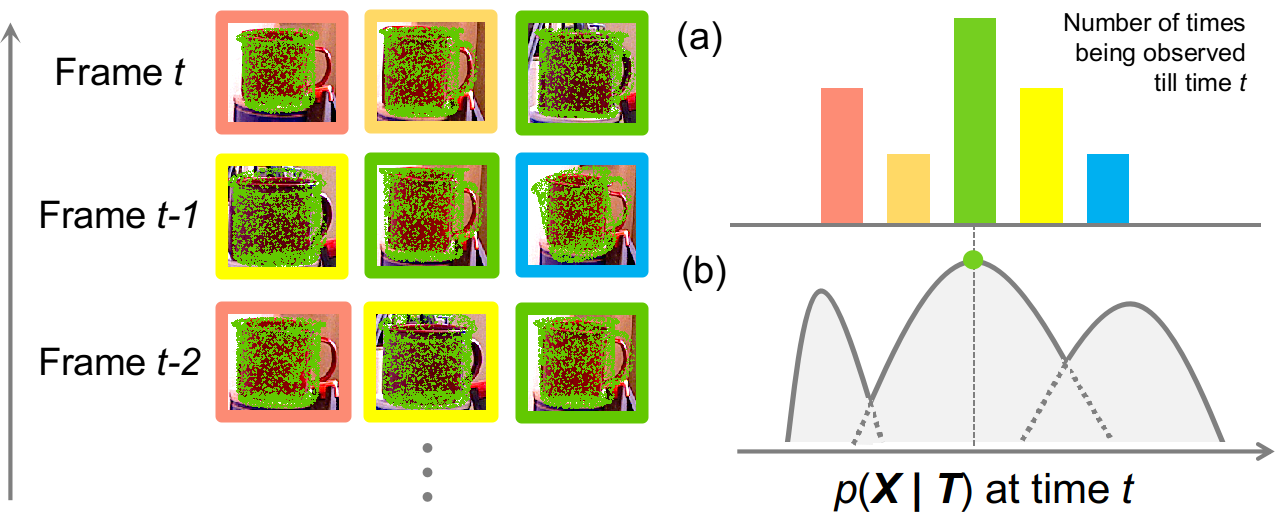}
        \caption{Max-mixtures for processing multiple pose hypotheses. Left half: current pool of pose hypotheses ($T_i$) for the landmark ``mug'' with same hypothesis in the same outline color till time $t$. (a) Number of times  each $T_i$  has been observed so far, which is positively correlated to the posterior distribution of robot/landmark poses, $p(\bf{X}|\bf{T})$, where $\textbf{T}$ denotes all multi-hypothesis measurements. Hence the significance of \emph{consistent inclusion} of the \emph{correct pose} within the hypothesis set. Otherwise max-mixtures may converge to a sub-optimal hypothesis. (b) Schematic probability distribution of three pose hypotheses with respect to $\textbf{X}$.}
    \label{mm}
\end{figure}

To efficiently solve the optimization problem under the multi-hypothesis formulation, we here adapt max-mixtures to model multi-modal measurements and implicitly seek out consistent landmark pose hypotheses via optimization. Previously, the max-mixture method has been leveraged to solve data association problems in SLAM~\cite{olson2013inference,doherty2020probabilistic} for its effectiveness in addressing large number of outliers in the least squares SLAM formulation. Thus by relying on max-mixtures' ability to distinguish the better-behaved hypothesis, the backend is able to recover a globally consistent set of robot/landmark poses out of an exponentially increasing number of ambiguous measurement combinations (see Fig.~\ref{mm}).

Generally, estimating poses from a sequence of observations can be formulated as a maximum likelihood estimation (MLE) problem:
\begin{equation}
\begin{split}
\hat{\mathbf{X}} &= \argmax_{\mathbf{X}}\prod _{i}{\phi(\mathbf{z}_i \mid \mathbf{X}_i)}
\end{split}
\label{eqn: MLE}
\end{equation}
where $\mathbf{X}_i$ denotes the latent variables involved in the $i^{th}$ input, $\mathbf{z}_i$ denotes the corresponding observation, and $\phi(\cdot)$ is the likelihood function for this observation. The set of all latent variables, $\mathbf{X}$, is the camera and object pose sequence being jointly optimized in the backend. For odometry measurements between robot poses, the additive measurement noise is modeled as the canonical Gaussian noise. For the object pose observation with $N$ hypotheses, a natural extension of the noise model is thus the sum of Gaussians or sum-mixtures:
\begin{equation}
{\phi(\mathbf{z}_i \mid \mathbf{X}_i)}= \sum_{j=1}^{N}{\omega_{ij}\mathcal{N}(\log(\mathbf{z}_{ij}h(\mathbf{X}_i)^{-1})^{\vee};\mathbf{0},\mathbf{\Sigma)}}
\label{eqn: sum}
\end{equation}
in which each component corresponds to a hypothesis and follows the convention of the perturbation model for 6D poses in \cite{barfoot2014associating}. With an assumption of equally probable hypothesis, the weight $\omega_{ij}$ for component $j$ is $\frac{1}{N}$. The measurement of interest here is the relative pose of the object with respect to the camera frame: $ ^{c} _{o}\mathbf{P} = h(\mathbf{X}_i) = h(^{w} _{o}\mathbf{P},\ ^{w} _{c}\mathbf{P}) = ^{c} _{w}\mathbf{P}   ^{w} _{o}\mathbf{P}$, which is a function of camera pose $^{w} _{c}\mathbf{P}$ and object pose $^{w} _{o}\mathbf{P}$ in the world frame.

Sum-mixtures falls outside the common nonlinear least squares (NLLS) optimization approaches for Gaussian noise models. For MLE estimation, we can instead consider the max-marginal\cite{koller2009probabilistic}, leading to the following max-mixture formulation:
\begin{equation}
    \phi(\mathbf{z}_i|\mathbf{X}_i) = \max_{j=1:N} {\omega_{ij}\mathcal{N}(\log(\mathbf{z}_{ij}h(\mathbf{X}_i)^{-1})^{\vee};\mathbf{0},\mathbf{\Sigma).}}\label{eqn: max mixture model}
\end{equation}

Max-mixtures switches to the ``max'' of all probability, retaining the better-behaved hypothesis while keeping the problem still within the realm of common NLLS optimization. Max-mixtures does not make a permanent choice out of multiple hypotheses. In an iteration of optimization, only one of the hypotheses actively contributes to the loss function and optimization steps via the ``max'' operator. Another hypothesis may turn active in the next iteration depending on the updated estimate of the latent variable, $\mathbf{X}_i$. Hence, by optimizing \eqref{eqn: MLE} for the updated estimate of $\mathbf{X}$, we evaluate all the Gaussian components in \eqref{eqn: max mixture model} and identify the most likely one as the hypothesis contributing to the estimate in the next optimization step. Please refer to \ref{implementation details} for implementation details about optimization.

\section{Experiments and Results}
In this section, in the goal of improving object-based SLAM results in the ambiguous environment setting, we would like to answer two questions: (1) Can MHPE predict a set of hypotheses that do contain the close-to-groundtruth object pose? (2) Can max-mixtures recover a globally consistent set of robot/landmark poses from MHPE measurements? To answer (1), we evaluate our approach on the YCB-Video benchmark in terms of distance loss and processing speed. For (2), we test our approach on a simulated video made with YCB objects for SLAM results, where the camera executes a much longer trajectory.

\subsection{Datasets}
\textbf{YCB-Video Dataset.} The YCB-Video Dataset~\cite{xiang2017posecnn} is built on 21 YCB objects\cite{calli2015ycb} placed in various indoor scenes and consists of 92 video sequences with 133,827 images in all. Each sequence provides RGB-D images, object segmentation masks, and annotated object poses. Following \cite{xiang2017posecnn}, we divide the whole dataset into 80 videos for training and select 2,949 keyframes from the rest of 12 videos for testing. An extra set of 80,000 synthetic images created by\cite{xiang2017posecnn} is also included in the training set.

 \begin{figure}
     \centering
     \includegraphics[scale=0.22]{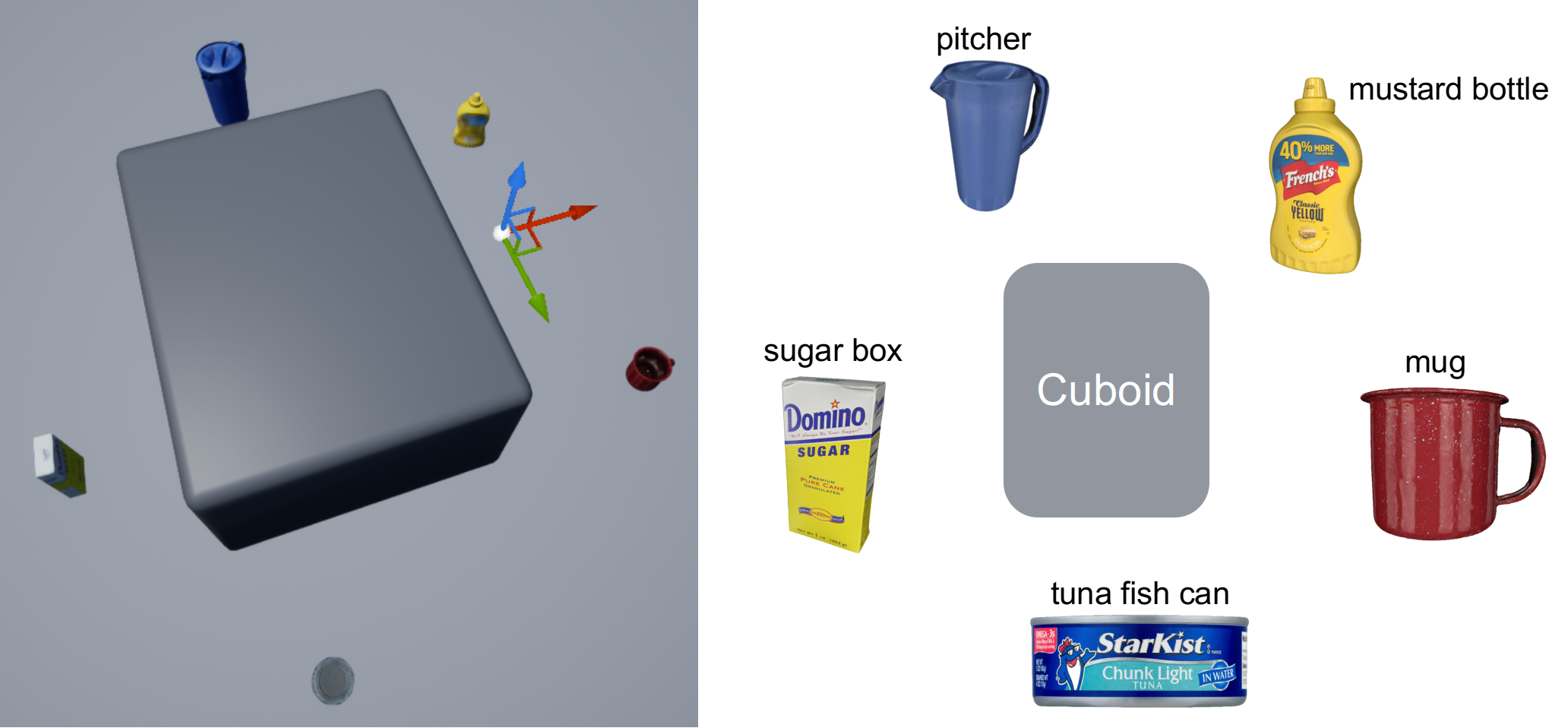}
     \caption{Simulated video topview and YCB object layout. Some YCB objects are deliberately chosen to bear more or less \emph{shape symmetries} (the mug and the tuna fish can), thus contributing to an \emph{ambiguous environment setting} for ground-level observations. Note that objects are also scaled up to 50 times of its original size so as to match with the size of the observer, i.e., the AirSim car simulator, in the video.}
     \label{airsim}
\end{figure}

\textbf{AirSim Simulated Video.}  Since the camera motion in the YCB-Video Dataset is relatively small compared to a typical SLAM-application scenario and its environment does not involve much observation ambiguity, we create a virtual environment of five YCB objects with some shape symmetries placed around a cuboid in Unreal Engine\cite{unrealengine} (see Fig.~\ref{airsim}). The AirSim\cite{shah2018airSim} car simulator is adopted to simulate a robot-mounted camera circling around the objects, thus making observations in an ambiguous environment setting. The observations provide RGB images, object segmentation masks, and automatically annotated 6D poses of the camera and the five objects. 

\subsection{Metrics}
To facilitate comparison within the YCB-Video Dataset benchmark, we follow two metrics used in \cite{xiang2017posecnn} and \cite{deng2019poserbpf}, i.e., ADD (introduced in \ref{learning obj}) and ADD-S. The ADD-S metric, i.e., average distance between the point in the estimation-transformed point set $\mathcal{X}_1$ and its closest point in the groundtruth-transformed point set $\mathcal{X}_2$, is defined as:
\begin{equation}
l_{ADD-S}=\frac{1}{|\mathcal{X}_1|}\sum_{\mathbf{x}_1\in\mathcal{X}_1}\underset{\mathbf{x}_2 \in \mathcal{X}_2}{\min}\|(\mathbf{\hat{R}}\mathbf{x}_1+\mathbf{\hat{t}})-(\mathbf{R}\mathbf{x}_2+\mathbf{t})\|\label{adds}
\end{equation}
where $\mathcal{X}_1$ and $\mathcal{X}_2$ are transformed from the object model point set.

The ADD metric reflects pose estimation error, while ADD-S focuses on shape similarity and is thus appropriate for evaluating symmetrical objects. Following \cite{xiang2017posecnn}, we set the distance threshold to 10cm and compute the area under the ADD and ADD-S curves (AUC) with higher AUC indicating higher accuracy.
\subsection{Implementation Details}\label{implementation details}
The MHPE network is implemented in PyTorch. We initialize ResNet-34 with weights pretrained on ImageNet. The final MLP module comprises of three fully connected layers of size 512, 512, and 256. The number of output hypotheses is set as $N$ = 5 for a good compromise between sufficient hypothesis diversity and additional computation overhead. For training, we use a batch size of 64 and train 100 epochs with the Adam optimizer. The initial learning rate is 0.0001 with a decay rate of 0.1 at epoch 50. The backend max-mixture algorithm is implemented in C++ using the Robot Operating System~\cite{quigley2009ros} and the iSAM2~\cite{kaess2012isam2} implementation from the GTSAM library\cite{dellaert2012factor}. All training and testing are conducted on a laptop with an Intel Core i7-9750H CPU and an Nvidia GeForce RTX 2070 GPU.

\subsection{Results on YCB-Video Dataset}
 We report our quantitative and qualitative object pose estimation results on the YCB-Video Dataset.
 
 Quantitatively, we compare our MHPE results to those of PoseCNN and the more recent PoseRBPF~\cite{deng2019poserbpf}, all with RGB images (see Table~\ref{tab1}) and object detection results from PoseCNN for the purpose of fair comparison. We choose the PoseRBPF-50-particle variant, which is of closer processing speed with ours and is therefore suitable for SLAM applications. 
 
 In particular, for MHPE results, we compare our hypothesis with the lowest ADD loss in each set (best hypothesis) against the single output from PoseCNN and PoseRBPF, as in our problem, it is the estimation error of the best hypothesis, rather than the average quality of the multi-hypothesis set, that indicates the existence of a consistent hypothesis mode for max-mixtures to switch to, and hence the possibility of recovering the true robot/landmark poses.

\begin{table*}[]
\centering
\caption{Quantitative ADD and ADD-S AUC Results on YCB-Video Dataset.
Symmetrical objects are in bold. Frame rate does not consider the effect of bounding box detection, which serves as part of pipeline inputs. Best ADD and ADD-S AUC values for each object are in bold (excluding ICP-refined results).}
\label{tab1}
\resizebox{\textwidth}{!}{%
\begin{tabular}{|l|c|c|c|c|c|c|c|c|}
\hline
\multirow{2}{*}{} & \multicolumn{6}{c|}{RGB} & \multicolumn{2}{c|}{RGB + Depth Image (for ICP)} \\ \cline{2-9} 
 & \multicolumn{2}{c|}{PoseCNN} & \multicolumn{2}{c|}{\begin{tabular}[c]{@{}c@{}}PoseRBPF\\ 50 particles\end{tabular}} & \multicolumn{2}{c|}{\begin{tabular}[c]{@{}c@{}}MHPE\\ Best Hypothesis\end{tabular}} & \multicolumn{2}{c|}{\begin{tabular}[c]{@{}c@{}}MHPE\\ Best Hypothesis + ICP\end{tabular}} \\ \hline
\multicolumn{1}{|c|}{\begin{tabular}[c]{@{}c@{}}Frame Rate\\ (fps)\end{tabular}} & \multicolumn{2}{c|}{5.9} & \multicolumn{2}{c|}{20.6} & \multicolumn{2}{c|}{32.3} & \multicolumn{2}{c|}{\begin{tabular}[c]{@{}c@{}}2.1\\ (CPU-only)\end{tabular}} \\ \hline
\multicolumn{1}{|c|}{Objects} & ADD & ADD-S & ADD & ADD-S & ADD & ADD-S & ADD & ADD-S \\ \hline
002\_master\_chef\_can & 50.9 & 84.0 & 56.1 & 75.6 & \textbf{67.9} & \textbf{93.8} & 74.3 & 90.0 \\ \hline
003\_cracker\_box & 51.7 & 76.9 & \textbf{73.4} & \textbf{85.2} & 67.8 & 82.9 & 86.8 & 91.6 \\ \hline
004\_sugar\_box & 68.6 & 84.3 & 73.9 & 86.5 & \textbf{83.1} & \textbf{91.3} & 97.4 & 98.2 \\ \hline
005\_tomato\_soup\_can & 66.0 & 80.9 & 71.1 & 82.0 & \textbf{79.5} & \textbf{92.2} & 38.1* & 66.9* \\ \hline
006\_mustard\_bottle & 79.9 & 90.2 & 80.0 & 90.1 & \textbf{81.6} & \textbf{90.8} & 98.0 & 98.7 \\ \hline
007\_tuna\_fish\_can & 70.4 & 87.9 & 56.1 & 73.8 & \textbf{78.0} & \textbf{92.5} & 87.0 & 95.8 \\ \hline
008\_pudding\_box & \textbf{62.9} & \textbf{79.0} & 54.8 & 69.2 & 45.4 & 71.5 & 83.7 & 92.7 \\ \hline
009\_gelatin\_box & 75.2 & 87.1 & \textbf{83.1} & \textbf{89.7} & 76.1 & 87.8 & 97.2 & 98.4 \\ \hline
010\_potted\_meat\_can & 59.6 & 78.5 & 47.0 & 61.3 & \textbf{69.1} & \textbf{85.5} & 66.4 & 78.4 \\ \hline
011\_banana & 72.3 & 85.9 & 22.8 & 64.1 & \textbf{87.7} & \textbf{93.7} & 93.8 & 97.3 \\ \hline
019\_pitcher\_base & 52.5 & 76.8 & 74.0 & 87.5 & \textbf{76.8} & \textbf{88.8} & 96.9 & 98.1 \\ \hline
021\_bleach\_cleanser & 50.5 & \textbf{71.9} & \textbf{51.6} & 66.7 & 47.7 & 70.3 & 88.8 & 94.1 \\ \hline
\textbf{024\_bowl} & 6.5 & 69.7 & 26.4 & \textbf{88.2} & \textbf{40.2} & 80.1 & 73.6 & 95.8 \\ \hline
025\_mug & 57.7 & 78.0 & \textbf{67.3} & \textbf{83.7} & 40.6 & 72.8 & 69.3 & 91.1 \\ \hline
035\_power\_drill & 55.1 & 72.8 & \textbf{64.4} & \textbf{80.6} & 39.5 & 71.2 & 61.1 & 81.7 \\ \hline
\textbf{036\_wood\_block} & 31.8 & 65.8 & 0.0 & 0.0 & \textbf{64.6} & \textbf{85.5} & 85.2 & 93.0 \\ \hline
037\_scissors & 35.8 & 56.2 & 20.6 & 30.9 & \textbf{64.5} & \textbf{88.9} & 63.5 & 88.9 \\ \hline
040\_large\_marker & 58.0 & 71.4 & 45.7 & 54.1 & \textbf{81.1} & \textbf{90.6} & 89.5 & 96.2 \\ \hline
\textbf{051\_large\_clamp} & 25.0 & 49.9 & 27.0 & \textbf{73.2} & \textbf{49.2} & 70.7 & 31.5 & 51.8 \\ \hline
\textbf{052\_extra\_large\_clamp} & 15.8 & 47.0 & \textbf{50.4} & \textbf{68.7} & 8.6 & 47.4 & 9.7 & 31.3 \\ \hline
\textbf{061\_foam\_brick} & 40.4 & 87.8 & \textbf{75.8} & 88.4 & 75.1 & \textbf{92.6} & 90.5 & 97.2 \\ \hline
All & 53.7 & 75.9 & 57.1 & 74.8 & \textbf{63.7} & \textbf{83.1} & 72.6 & 85.2 \\ \hline
\end{tabular}%
}
\end{table*}

\begin{figure}
    \centering
    \includegraphics[width=\linewidth]{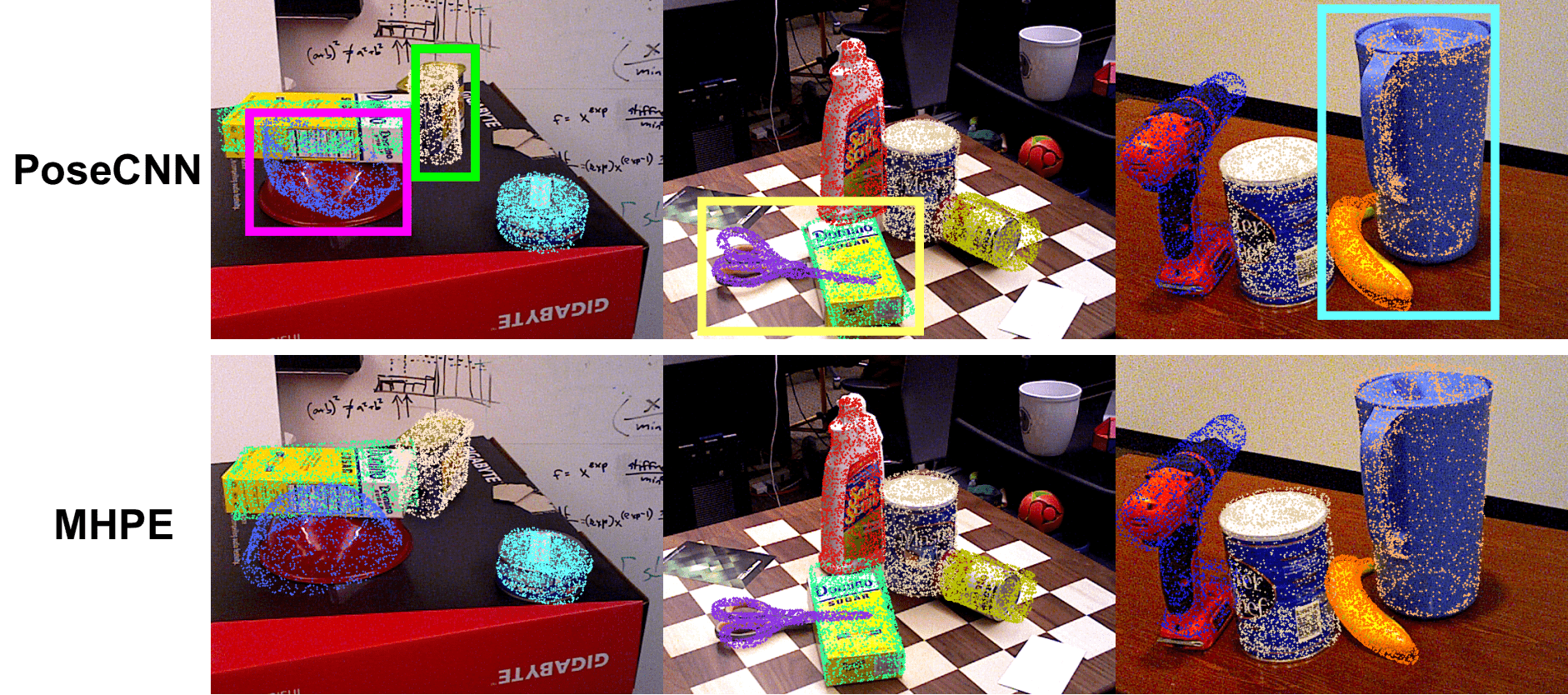}
    \caption{Qualitative results of object pose estimation. Each object pose is annotated with the 2D image frame projection of a colored point cloud transformed by the pose estimate. Improved areas are marked by colored bounding boxes in the first row showing PoseCNN results.}
    \label{compare}
\end{figure}

    Qualitatively, we display some sample visualizations of the object pose estimate in Fig.~\ref{compare} for comparison between PoseCNN and the best hypothesis of our MHPE results. Since PoseRBPF does not provide multi-object pose estimation results on the YCB-Video Dataset, we do not include them in Fig.~\ref{compare}.

\textbf{Overall Performance.} Table~\ref{tab1} presents the ADD and ADD-S AUC values of all the 21 objects in the YCB-Video Dataset and the rough processing speed of each approach. We can conclude that with similar processing speed (ours: 23.0 fps \& poseRBPF: 20.6 fps, excluding object detection as its results are used as system inputs), our best hypothesis from the MHPE frontend outperforms PoseCNN and PoseRBPF by 18.6\%,  11.6\% on ADD, and 9.4\%, 11.1\% on ADD-S, respectively. This demonstrates the effectiveness of our multi-hypothesis strategy in improving pose estimation accuracy and the reliability in providing valid hypothesis set for supporting the backend optimization of robot/landmark poses.

As shown by Fig.~\ref{compare}, when compared to PoseCNN, our best hypotheses from MHPE give more accurate pose estimations for the upside-down bowl in the first scene, the scissors and sugar box in the second, and the banana and the blue pitcher in the last scene.

\textbf{Dealing with Ambiguity.} From Table~\ref{tab1}, it is observed that the MHPE network performs better on most of the symmetrical objects (marked in bold). This proves the effectiveness of our object pose estimation strategy in dealing with shape ambiguity, as it is easier for symmetrical objects to have ambiguous poses, which may induce the network to generate random false positive hypotheses. 

We further explain this with the ``clamp'' example in Fig.~\ref{variety}. The five pose hypotheses rendered here are MHPE's estimates for the clamp pose, which include the correct estimate, (e), in green. Considering the diversity of the hypothesis set, which consists of  seemingly  inaccurate,  and  mirror-image  pair hypotheses, a single-output network may throw out a random estimate as any of (a) \& (d)  or (b) \& (e), since these two mirror-image pairs tend to incur similar ADD loss during training with respect to the grey mirror plane (shown in (f)). Therefore, with each output branch evolving towards different domains of the solution space, MHPE can first produce a set of possible pose candidates and then let the backend to pick one based on the observations accumulated so far. The general poor performance on the extra large clamp may result from the given bounding box quality as it is hard for PoseCNN to distinguish between ``large clamp'' and ``extra large clamp''. We also argue that for objects with poorer MHPE performance, these object pose hypotheses are consistently less accurate instead of randomly varying, and is hence less likely to 
cause big jumps in the backend robot pose estimation. Moreover, those jumps should be an even rarer occurrence considering the smoothing effect brought by the accumulation of observations in the backend.

\textbf{ICP Post-Processing.} We also provide AUC results with ICP post-processing, where the refined poses achieve significant accuracy improvements. While we are not comparing them to the non-post-processing PoseCNN and PoseRBPF results here, considering the great improvement in AUC values after ICP, we argue that our MHPE-predicted hypotheses are on average not far from the groundtruth and thus effectively prevent ICP from local-minimum failure. The only exception is for the ``tomato soup can'' (marked with asterisks) as in some frames, the lack of enough visible points from heavy occlusion greatly impairs the registration quality. 

\subsection{SLAM Results for the Simulated Video}
We apply our approach to the simulated video and test its effectiveness in the scenario with larger camera motion. For evaluation purposes, we compare the robot/landmark pose sequence optimized from our MHPE+max-mixtures approach against those obtained by two practical approaches when dealing with multiple hypotheses: averaging and random selection. Additionally, the random selection approach also resembles the behavior of the traditional single-hypothesis approach under ambiguous settings, as the frontend tends to throw out a random time-inconsistent measurement.

\begin{figure}
    \centering
    \includegraphics[width=\linewidth]{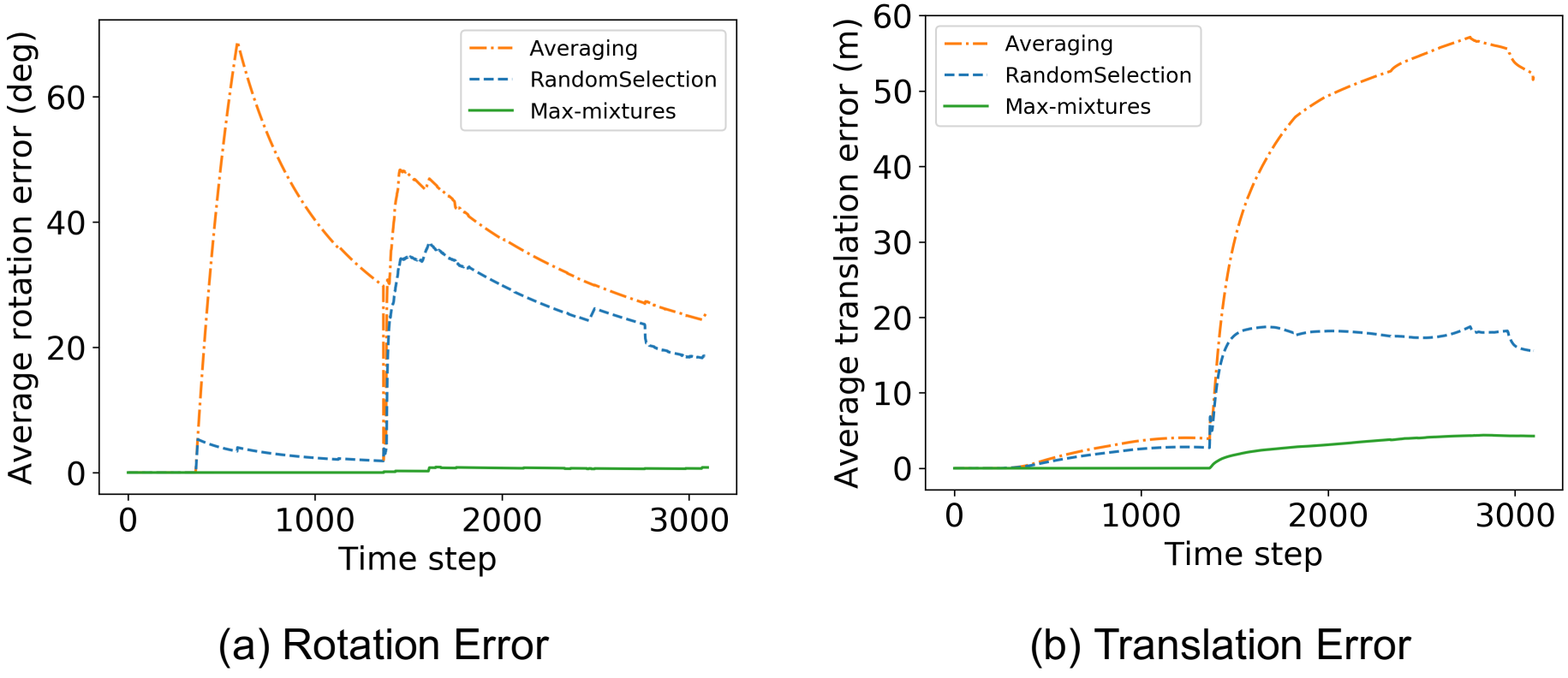}
    \caption{Running average of the estimation error for rotation and translation across the course the robot motion. The groundtruth is extracted from the simulator.}
    \label{time}
\end{figure} 

To better adapt to the simulated environment, in addition to the YCB-Video data, the MHPE network is trained on an extra of 18,523 images of a robot circumnavigating around each object at different ranges. The testing is then run on a 3,100-frame sequence in a similar environment, with the camera moving first in the inner and then the outer circle of the area enclosed by the five objects. Objects are detected in 1,611 frames of the sequence and the resulting pose estimations are used as the input to the SLAM backend.

\begin{figure}
    \centering
    \includegraphics[scale=0.35]{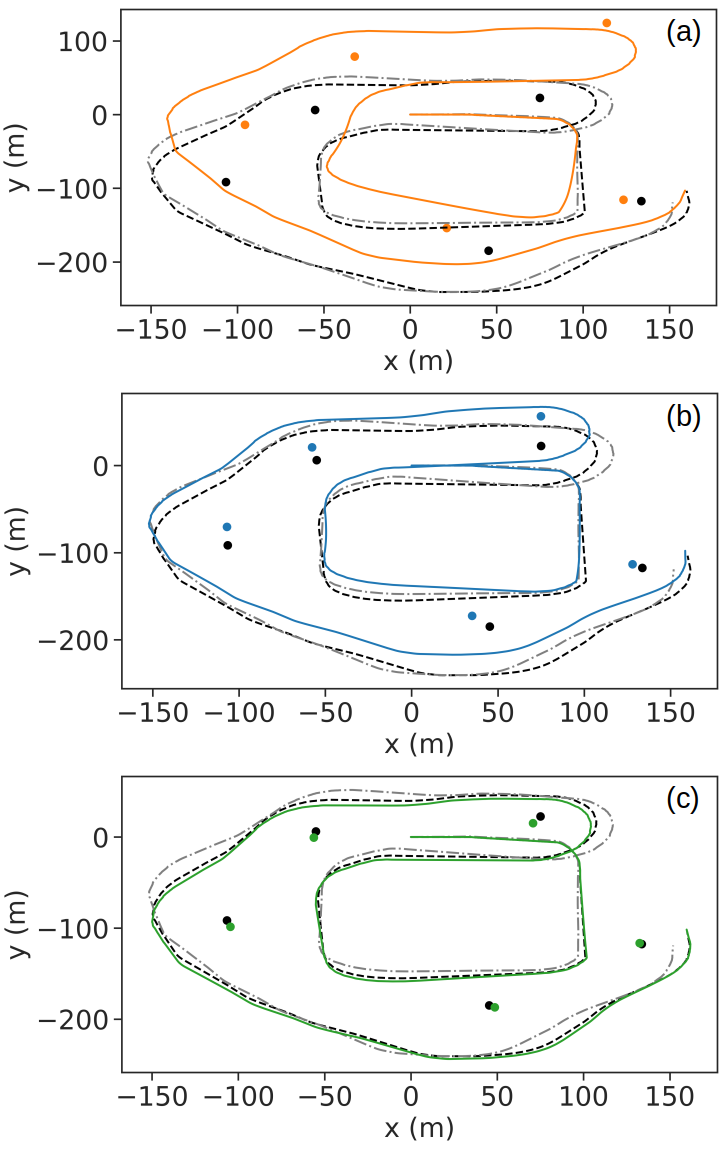}
    \caption{Comparison of camera trajectories and object position estimates. We use evo~\cite{grupp2017evo} for trajectory evaluation. Objects are shown as dots. Groundtruth is shown in black. Odometry trajectories are shown by dash-dotted grey lines. (a) Averaging; (b) Random Selection; (c) Max-mixtures.}
    \label{traj}
\end{figure}

 \textbf{Robot Pose Estimation.} Quantitatively, in Fig.~\ref{time}, we plot the running average of the estimation error for rotation and translation, respectively. As shown in Fig.~\ref{time}, we can conclude that generally max-mixtures performs the best while averaging does the worst. The discontinuities in (a) corresponds to the emergence of new objects. Considering the ambiguous measurement from the deliberately selected symmetrical objects, with a continuously lower rotation/translation error, our approach shows its robustness to the disturbance of the false-positive pose hypotheses and outperforms random selection, which to some extent, also indicates how a common, single-hypothesis SLAM system will behave. In Fig.~\ref{time} (b), the sudden climb in the translation error at about halfway through the course can be attributed to the reappearance of the mug in the video, and averaging obviously suffers from considerable drift due to ambiguous measurements from this loop closure.
 
 Qualitatively, it can be observed from Fig.~\ref{traj} that in terms of estimation accuracy, our approach outperforms both averaging and random selection by a large margin. We argue that by leveraging the accumulated observation from previous frames, our approach is able to  recover consistent estimates from the ambiguous pose measurements, as opposed to its single-hypothesis-based counterparts.  
 
 This is especially useful for treating shape-ambiguous objects and loop closures. For example, the camera could obtain a better pose estimate of the mug by taking into account when last time the mug handle was visible, and this will, in turn, benefit loop closure detection. Furthermore, the result also implies that averaging over the candidate pool is the least effective approach, as the result of averaging over rotation matrices from the hypothesis ensemble can deviate from the actual rotation of any member hypothesis~\cite{hartley2013rotation}.

\begin{figure}
    \centering
    \includegraphics[width=\linewidth]{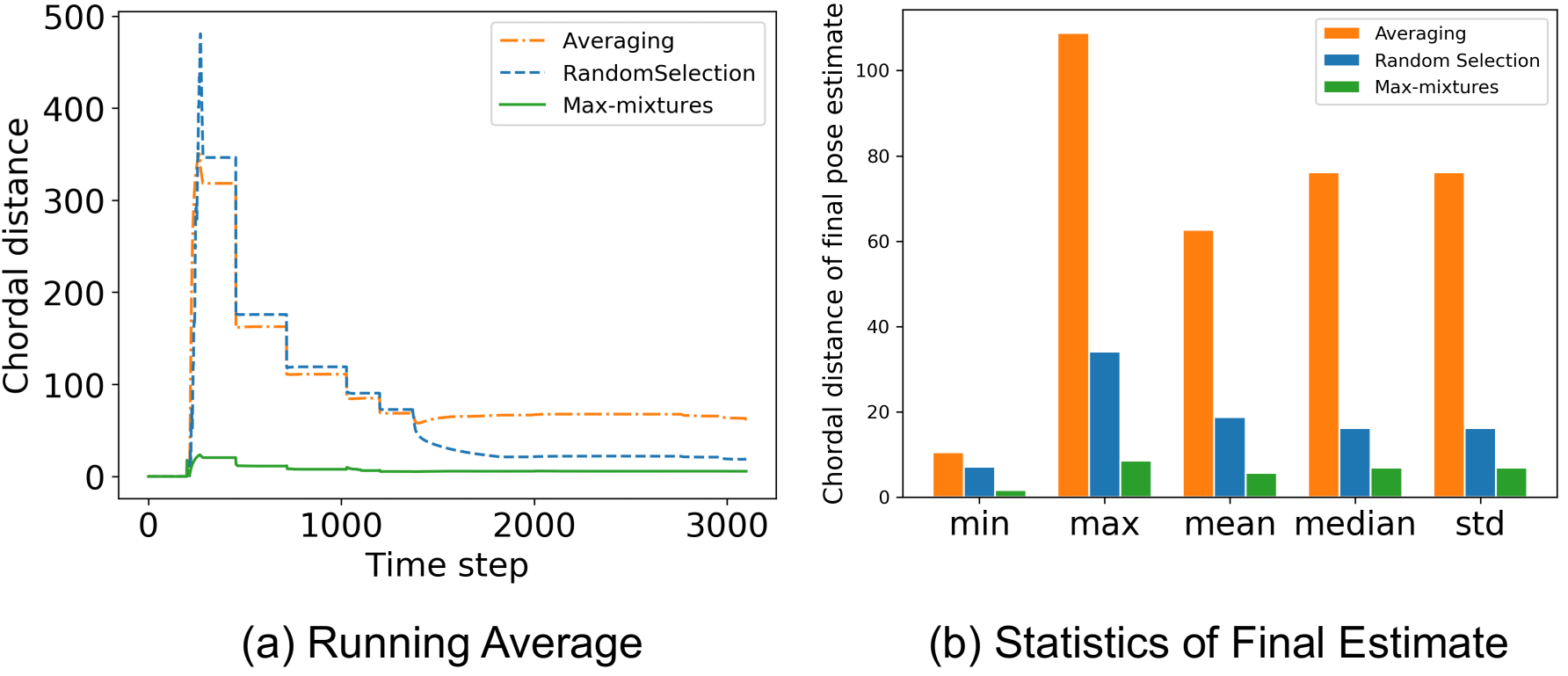}
    \caption{Landmark pose estimation error.}
    \label{lmk_chord}
\end{figure}

\textbf{Landmark Pose Estimation.} Here, we choose landmark pose estimation from MHPE as the frontend measurement to help recover robot poses. We compute the chordal distance between the groundtruth and the estimated landmark pose for comparison (see Fig.~\ref{lmk_chord}).

As shown in Fig.~\ref{lmk_chord} (a), we could observe the pattern of a step-wise increase in error each time a new landmark is observed, which is then followed by a gradual error decrease as optimization proceeds.  Compared to the other two approaches, our approach exhibits a much less steep surge between ``stairs'' as though with no prior knowledge about the object, it could employ past knowledge to help gradually switch to the better-behaved mode. Around time step 1,400 when the robot finishes its first circle and returns back to the mug, while averaging merges ambiguous measurements at the loop closure and hence renders the optimization off track (corresponds to trajectory in Fig.~\ref{traj} (a)), max-mixtures manages to maintain a lower average error. 

The above behaviors are also reflected in Fig.~\ref{lmk_chord} (b), as for the final estimate of object poses, max-mixtures proves its superiority in all statistical descriptions, indicating a constantly higher estimation precision amongst  all the three methods.

\section{Conclusion}
In this paper we develop a novel approach that  allows the robot to reason  about  multiple  hypotheses  for  an  object’s  pose,  and synthesize  this  locally  ambiguous  information  into  a  globally consistent  object  and  robot  trajectory  estimation  formulation. We instantiate this approach by designing a deep-learning-based frontend for producing multiple object pose hypotheses and a max-mixture-based backend for selecting pose candidates and conducting joint robot/landmark pose optimization. Experiments on the real YCB-Video Dataset and the synthetic YCB object video demonstrate the effectiveness of our approach to performing object pose and robot trajectory estimation in ambiguous environment settings. Future work may involve providing extra clues to facilitate max-mixtures optimization, such as by regressing weights for each pose hypothesis from the MHPE frontend.

\bibliographystyle{ieeetr}
\bibliography{ref} 

\begin{thebibliography}{10}

\bibitem{rosen2021advances}
D.~M. Rosen, K.~J. Doherty, A.~T. Espinoza, and J.~J. Leonard, ``Advances in
  {I}nference and {R}epresentation for {S}imultaneous {L}ocalization and
  {M}apping,'' {\em Annual Review of Control, Robotics, and Autonomous
  Systems}, vol.~4, 2021.

\bibitem{salas2013slam++}
R.~F. Salas-Moreno, R.~A. Newcombe, H.~Strasdat, P.~H. Kelly, and A.~J.
  Davison, ``{SLAM}++: Simultaneous localisation and mapping at the level of
  objects,'' in {\em Proceedings of the IEEE conference on computer vision and
  pattern recognition}, pp.~1352--1359, 2013.

\bibitem{olson2013inference}
E.~Olson and P.~Agarwal, ``Inference on networks of mixtures for robust robot
  mapping,'' {\em The International Journal of Robotics Research}, vol.~32,
  no.~7, pp.~826--840, 2013.

\bibitem{collet2011moped}
A.~Collet, M.~Martinez, and S.~S. Srinivasa, ``The moped framework: Object
  recognition and pose estimation for manipulation,'' {\em The international
  journal of robotics research}, vol.~30, no.~10, pp.~1284--1306, 2011.

\bibitem{zhu2014single}
M.~Zhu, K.~G. Derpanis, Y.~Yang, S.~Brahmbhatt, M.~Zhang, C.~Phillips,
  M.~Lecce, and K.~Daniilidis, ``Single image 3d object detection and pose
  estimation for grasping,'' in {\em 2014 IEEE International Conference on
  Robotics and Automation (ICRA)}, pp.~3936--3943, IEEE, 2014.

\bibitem{aubry2014seeing}
M.~Aubry, D.~Maturana, A.~A. Efros, B.~C. Russell, and J.~Sivic, ``Seeing 3d
  chairs: exemplar part-based 2d-3d alignment using a large dataset of cad
  models,'' in {\em Proceedings of the IEEE conference on computer vision and
  pattern recognition}, pp.~3762--3769, 2014.

\bibitem{xiang2017posecnn}
Y.~Xiang, T.~Schmidt, V.~Narayanan, and D.~Fox, ``{PoseCNN}: A convolutional
  neural network for {6D}object pose estimation in cluttered scenes,'' {\em
  arXiv preprint arXiv:1711.00199}, 2017.

\bibitem{kehl2017ssd6d}
W.~Kehl, F.~Manhardt, F.~Tombari, S.~Ilic, and N.~Navab, ``{SSD-6D}: Making
  {RGB}-based {3D} detection and {6D} pose estimation great again,'' in {\em
  IEEE Conference on Computer Vision and Pattern Recognition (CVPR)}, 2017.

\bibitem{tremblay2018deep}
J.~Tremblay, T.~To, B.~Sundaralingam, Y.~Xiang, D.~Fox, and S.~Birchfield,
  ``Deep object pose estimation for semantic robotic grasping of household
  objects,'' {\em arXiv preprint arXiv:1809.10790}, 2018.

\bibitem{wang2019densefusion}
C.~Wang, D.~Xu, Y.~Zhu, R.~Mart{\'\i}n-Mart{\'\i}n, C.~Lu, L.~Fei-Fei, and
  S.~Savarese, ``{DenseFusion}: {6D} object pose estimation by iterative dense
  fusion,'' in {\em Proceedings of the IEEE Conference on Computer Vision and
  Pattern Recognition}, pp.~3343--3352, 2019.

\bibitem{Besl1992}
P.~J. Besl and N.~D. McKay, ``A method for registration of {3-D} shapes,'' {\em
  IEEE Transactions on Pattern Analysis and Machine Intelligence}, vol.~14,
  pp.~239--256, Feb. 1992.

\bibitem{Hahnel02probabilisticmatching}
D.~Hähnel and W.~Burgard, ``Probabilistic matching for {3D} scan
  registration,'' in {\em Proceedings of the VDI Conference}, 2002.

\bibitem{agamennoniIROS16}
G.~Agamennoni, S.~Fontana, R.~Siegwart, and D.~Sorrenti, ``Point clouds
  registration with probabilistic data association,'' in {\em Proceedings of
  The International Conference on Intelligent Robots and Systems (IROS)}, 2016.

\bibitem{iser_2016_hinzmann}
T.~Hinzmann, T.~Stastny, G.~Conte, P.~Doherty, P.~Rudol, M.~Wzorek,
  E.~Galceran, R.~Siegwart, and I.~Gilitschenski, ``Collaborative {3D}
  reconstruction using heterogeneous uavs: System and experiments,'' in {\em
  International Symposium on Experimental Robotics}, 2016.

\bibitem{Wong_2017}
J.~M. Wong, V.~Kee, T.~Le, S.~Wagner, G.-L. Mariottini, A.~Schneider,
  L.~Hamilton, R.~Chipalkatty, M.~Hebert, D.~M. Johnson, and et~al.,
  ``Seg{ICP}: Integrated deep semantic segmentation and pose estimation,'' {\em
  2017 IEEE/RSJ International Conference on Intelligent Robots and Systems
  (IROS)}, Sep 2017.

\bibitem{Wang_2019_ICCV}
Y.~Wang and J.~M. Solomon, ``Deep closest point: Learning representations for
  point cloud registration,'' in {\em The IEEE International Conference on
  Computer Vision (ICCV)}, October 2019.

\bibitem{Wang_2019_NeurIPS}
Y.~Wang and J.~M. Solomon, ``{PRNet}: Self-supervised learning for
  partial-to-partial registration,'' in {\em 33rd Conference on Neural
  Information Processing Systems}, 2019.

\bibitem{1241885}
M.~{Montemerlo} and S.~{Thrun}, ``Simultaneous localization and mapping with
  unknown data association using {FastSLAM},'' in {\em 2003 IEEE International
  Conference on Robotics and Automation (Cat. No.03CH37422)}, vol.~2,
  pp.~1985--1991 vol.2, 2003.

\bibitem{sunderhauf2012towards}
N.~S{\"u}nderhauf and P.~Protzel, ``Towards a robust back-end for pose graph
  {SLAM},'' in {\em 2012 IEEE international conference on robotics and
  automation}, pp.~1254--1261, IEEE, 2012.

\bibitem{6385590}
N.~{Sünderhauf} and P.~{Protzel}, ``Switchable constraints for robust pose
  graph {SLAM},'' in {\em 2012 IEEE/RSJ International Conference on Intelligent
  Robots and Systems}, pp.~1879--1884, 2012.

\bibitem{doherty2020probabilistic}
K.~J. Doherty, D.~P. Baxter, E.~Schneeweiss, and J.~J. Leonard, ``Probabilistic
  data association via mixture models for robust semantic {SLAM},'' in {\em
  2020 IEEE International Conference on Robotics and Automation (ICRA)},
  pp.~1098--1104, IEEE, 2020.

\bibitem{nicholson2018quadricslam}
L.~Nicholson, M.~Milford, and N.~S{\"u}nderhauf, ``Quadric{SLAM}: Dual quadrics
  from object detections as landmarks in object-oriented {SLAM},'' {\em IEEE
  Robotics and Automation Letters}, vol.~4, no.~1, pp.~1--8, 2018.

\bibitem{yang2019cubeslam}
S.~Yang and S.~Scherer, ``Cube{SLAM}: Monocular 3-{D} object {SLAM},'' {\em
  IEEE Transactions on Robotics}, vol.~35, no.~4, pp.~925--938, 2019.

\bibitem{deng2019poserbpf}
X.~Deng, A.~Mousavian, Y.~Xiang, F.~Xia, T.~Bretl, and D.~Fox, ``Poserbpf: A
  rao-blackwellized particle filter for 6d object pose tracking,'' {\em arXiv
  preprint arXiv:1905.09304}, 2019.

\bibitem{he2016deep}
K.~He, X.~Zhang, S.~Ren, and J.~Sun, ``Deep residual learning for image
  recognition,'' in {\em Proceedings of the IEEE conference on computer vision
  and pattern recognition}, pp.~770--778, 2016.

\bibitem{he2017mask}
K.~He, G.~Gkioxari, P.~Doll{\'a}r, and R.~Girshick, ``Mask {R-CNN},'' in {\em
  Proceedings of the IEEE international conference on computer vision},
  pp.~2961--2969, 2017.

\bibitem{hinterstoisser2012model}
S.~Hinterstoisser, V.~Lepetit, S.~Ilic, S.~Holzer, G.~Bradski, K.~Konolige, and
  N.~Navab, ``Model based training, detection and pose estimation of
  texture-less {3D} objects in heavily cluttered scenes,'' in {\em Asian
  conference on computer vision}, pp.~548--562, Springer, 2012.

\bibitem{guzman2012multiple}
A.~Guzman-Rivera, D.~Batra, and P.~Kohli, ``Multiple choice learning: Learning
  to produce multiple structured outputs,'' in {\em Advances in Neural
  Information Processing Systems}, pp.~1799--1807, 2012.

\bibitem{barfoot2014associating}
T.~D. Barfoot and P.~T. Furgale, ``Associating uncertainty with
  three-dimensional poses for use in estimation problems,'' {\em IEEE
  Transactions on Robotics}, vol.~30, no.~3, pp.~679--693, 2014.

\bibitem{koller2009probabilistic}
D.~Koller and N.~Friedman, {\em Probabilistic graphical models: principles and
  techniques}.
\newblock MIT press, 2009.

\bibitem{calli2015ycb}
B.~Calli, A.~Singh, A.~Walsman, S.~Srinivasa, P.~Abbeel, and A.~M. Dollar,
  ``The {YCB} object and model set: Towards common benchmarks for manipulation
  research,'' in {\em 2015 international conference on advanced robotics
  (ICAR)}, pp.~510--517, IEEE, 2015.

\bibitem{unrealengine}
{Epic Games}, ``Unreal engine.''

\bibitem{shah2018airSim}
S.~Shah, D.~Dey, C.~Lovett, and A.~Kapoor, ``{AirSim}: High-fidelity visual and
  physical simulation for autonomous vehicles,'' in {\em Field and service
  robotics}, pp.~621--635, Springer, 2018.

\bibitem{quigley2009ros}
M.~Quigley, K.~Conley, B.~Gerkey, J.~Faust, T.~Foote, J.~Leibs, R.~Wheeler, and
  A.~Y. Ng, ``{ROS}: an open-source robot operating system,'' in {\em ICRA
  workshop on open source software}, vol.~3, p.~5, Kobe, Japan, 2009.

\bibitem{kaess2012isam2}
M.~Kaess, H.~Johannsson, R.~Roberts, V.~Ila, J.~J. Leonard, and F.~Dellaert,
  ``isam2: Incremental smoothing and mapping using the bayes tree,'' {\em The
  International Journal of Robotics Research}, vol.~31, no.~2, pp.~216--235,
  2012.

\bibitem{dellaert2012factor}
F.~Dellaert, ``Factor graphs and {GTSAM}: A hands-on introduction,'' tech.
  rep., Georgia Institute of Technology, 2012.

\bibitem{grupp2017evo}
M.~Grupp, ``evo: Python package for the evaluation of odometry and {SLAM}..''
  \url{https://github.com/MichaelGrupp/evo}, 2017.

\bibitem{hartley2013rotation}
R.~Hartley, J.~Trumpf, Y.~Dai, and H.~Li, ``Rotation averaging,'' {\em
  International journal of computer vision}, vol.~103, no.~3, pp.~267--305,
  2013.

\end{thebibliography}
\end{document}